\documentclass[10pt]{article} 
\usepackage[preprint]{tmlr}

\usepackage{amsmath,amsfonts,bm}

\def\eqref#1{equation~\ref{#1}}

\def\1{\bm{1}}

\DeclareMathAlphabet{\mathsfit}{\encodingdefault}{\sfdefault}{m}{sl}
\SetMathAlphabet{\mathsfit}{bold}{\encodingdefault}{\sfdefault}{bx}{n}

\DeclareMathOperator*{\argmin}{arg\,min}

\usepackage{hyperref}
\usepackage{url}
\usepackage{dirtytalk}
\usepackage{layouts}
\usepackage{multirow}
\usepackage{rotating}
\usepackage[ruled]{algorithm2e}
\usepackage{makecell}
\usepackage{placeins}
\usepackage{wrapfig}
\usepackage{color,soul}
\usepackage[dash,dot]{dashundergaps}
\usepackage{cleveref}
\usepackage{booktabs}
\usepackage{xspace}

\newcommand{\std}[1]{\textcolor[gray]{0.3}{\scriptsize$\pm$#1}}

\definecolor{Gray}{gray}{0.9}
\definecolor{DarkGray}{gray}{0.5}
\definecolor{greenish}{HTML}{55a868}
\definecolor{yellowish}{HTML}{FFD92F}
\definecolor{seaborn_blue}{HTML}{5471AB}
\definecolor{seaborn_orange}{HTML}{D1885C}
\definecolor{seaborn_green}{HTML}{6AA66E}
\definecolor{seaborn_red}{HTML}{B65655}

\title{How Low Can You Go? Identifying Prototypical In-Distribution Samples for Unsupervised Anomaly Detection}

\author{\name Felix Meissen \email felix.meissen@tum.de\\
      \addr Technical University of Munich
      \AND
      \name Johannes Getzner\\
      \addr Technical University of Munich
      \AND
      \name Alexander Ziller\\
      \addr Technical University of Munich
      \AND
      \name Özgün Turgut\\
      \addr Technical University of Munich
      \AND
      \name Georgios Kaissis\\
      \addr Technical University of Munich
      \AND
      \name Martin J. Menten\\
      \addr Technical University of Munich
      \AND
      \name Daniel Rueckert\\
      \addr Technical University of Munich}

\def\month{MM}  
\def\year{YYYY} 
\def\openreview{\url{https://openreview.net/forum?id=XXXX}} 

\begin{document}

\maketitle

\begin{abstract}
Unsupervised anomaly detection (UAD) alleviates large labeling efforts by training exclusively on unlabeled in-distribution data and detecting outliers as anomalies.
Generally, the assumption prevails that large training datasets allow the training of higher-performing UAD models.
However, in this work, we show that UAD with extremely few training samples can already match -- and in some cases even surpass -- the performance of training with the whole training dataset.
Building upon this finding, we propose an unsupervised method to reliably identify prototypical samples to further boost UAD performance.
We demonstrate the utility of our method on seven different established UAD benchmarks from computer vision, industrial defect detection, and medicine.
With just 25 selected samples, we even exceed the performance of full training in $25/67$ categories in these benchmarks.
Additionally, we show that the prototypical in-distribution samples identified by our proposed method generalize well across models and datasets and that observing their sample selection criteria allows for a successful manual selection of small subsets of high-performing samples.
Our code is available at \url{https://anonymous.4open.science/r/uad_prototypical_samples/}
\end{abstract}

\section{Introduction}
\label{sec:introduction}

Unsupervised anomaly detection (UAD) or out-of-distribution (OOD) detection aims to distinguish samples from an in-distribution (ID) from any sample that stems from another distribution.
To address this task, typically, machine-learning models are employed to represent the in-distribution by exclusively using samples from that distribution for training.
The converged model detects OOD samples via their distance to the in-distribution.
Compared to supervised training, this setup alleviates the need for large labeled datasets, is not susceptible to class imbalance, and is not restricted to anomalies seen during training.
Due to these advantages, UAD has several vital applications in computer vision: It is used to detect pathological samples in medical images \citep{fanogan,lagogiannis_2023,bercea2022federated,fae}, to spot defects in industrial manufacturing \citep{mvtecad,patchcore,rd,pni}, or as safeguards to filter unsuitable input data for supervised downstream models, for example, in autonomous driving.

\begin{figure}[t]
    \centering
    \includegraphics[width=\linewidth]{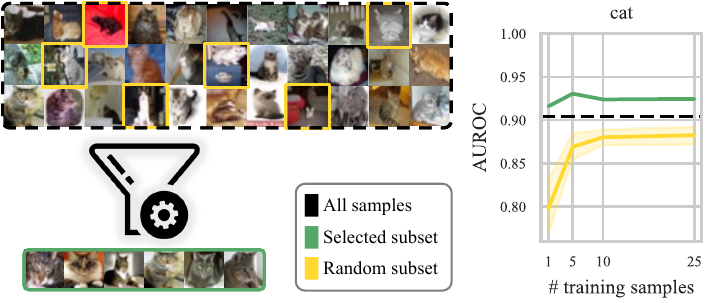}
    \caption{Selecting only a few prototypical in-distribution samples (identified by our method) for training can result in higher anomaly detection performance than training with $100\%$ of the available data. Results for anomaly detection on the \textit{cat} class from \mbox{CIFAR10}. Black dashed: \dashuline{full training}. Yellow: \setulcolor{yellowish}\ul{randomly selected samples}, including standard deviations over different random selections. Green: \setulcolor{greenish}\ul{Best-performing samples identified with our method}.}
    \label{fig:teaser_figure}
\end{figure}

In deep learning, the prevailing assumption is that more data leads to better models.
%
%
However, training with only very few samples would, have numerous advantages.
Small datasets are cheap, easy to obtain, and also available for a wider variety of tasks.
Additionally, it also makes more tasks feasible to solve where data may be very difficult or expensive to acquire.
Moreover, small training datasets lead to better explainability since output scores can directly be related to (dis-)similarities to the training data.
This improved explainability, in turn, lowers the entry bar for designing performant algorithms, especially for actors with few resources, thus contributing to the democratization of AI.
%
%
While overfitting with very few training samples diminishes the utility of supervised machine learning algorithms, UAD models are not impacted in the same way.
In fact, they rely on overfitting to the ID data and, as a result, not generalizing to OOD data.
This makes UAD suitable for training with extremely small datasets.

In this paper, we present findings showing that only very few training samples are required to achieve similar or even better anomaly detection performance compared to training with $100\%$ of the available training data (which we denote as \say{full training} in the remainder of the manuscript).
Through the unification of concepts from anomaly detection and core-set selection, we additionally propose an unsupervised method for selecting a high-performing subset of samples from the initial training dataset, as depicted in \Cref{fig:teaser_figure}, and evaluate the effectiveness of this approach on a multitude of different models, datasets, and tasks.
We further propose two weakly supervised selection strategies that serve as weak upper bounds for the introduced unsupervised sample selection and provide additional insights into the method's selection process.
%
In summary, the main contributions of this paper are:
\begin{itemize}
    \item We show for the first time that an exceedingly small number of training samples can suffice for performant, robust, and interpretable UAD, achieving state-of-the-art performance on a multitude of established benchmarks.
    \item We propose an unsupervised method to reliably find well-performing subsets prototypical in-distribution samples and describe their common characteristics.
    \item We further demonstrate that the prototypical samples identified by our method and their characteristics translate to equally good performance for other models, datasets, and even tasks.
    \item Lastly, we provide a theoretical justification explaining the increase in performance through training with very few samples.
\end{itemize}

\section{Related Work}\label{sec:related_work}
Our work combines ideas from both the fields of anomaly detection and core-set selection.  
Here, we give a brief overview of these concepts and related research work.

\subsection{Anomaly Detection}
Anomaly detection is deeply rooted in computer vision, with many influential works benchmarking their models on natural-image datasets, such as CIFAR10 and CIFAR100 \citep{cifar10}, MNIST \citep{mnist}, or Fashion-MNIST \citep{fashion_mnist}.
Early works attempting to solve UAD on these benchmarks were mostly based on (variational) autoencoders \citep{zhou2017anomaly,kim2019rapp,liu2020towards,abati2019latent} or GANs \citep{perera2019ocgan,deecke2019image,akcay2019ganomaly} trying to restrict the learned manifold of the generative model.
The model is expected to faithfully reconstruct in-distribution samples, whereas OOD samples can be detected due to their large reconstruction errors.
Also, one-class classification models \citep{deepsvdd} or ones that learn surrogate tasks \citep{golan2018deep,bergman2020classification} have been successfully used.
More recently, works based on pre-trained neural networks (such as ResNets \citet{resnet}) have become popular and still are the best-performing models for the aforementioned datasets \citep{bergman2020deep}.

The release of MVTec-AD \citep{mvtecad} for industrial defect detection sparked several works focusing on this dataset as it was the first to contain a variety of useful, real-world anomaly detection tasks.
After early attempts to solve these with various techniques, including autoencoders \citep{bergmann2018improving} and knowledge distillation methods \citep{bergmann2020uninformed}, research converged on self-supervised approaches \citep{draem,cutpaste}, ResNets pre-trained on ImageNet \citep{padim,rd,patchcore}, or combinations thereof \citep{pni}.

Anomaly detection was also successfully applied in medical computer vision, where it is used to discriminate samples from healthy subjects (in-distribution) from diseased ones (outliers). \citet{fanogan} have successfully discovered biomarkers in retinal OCT images using a GAN.
To detect tumors and lesions in brain MRI, numerous autoencoder-based approaches \citep{you2019unsupervised,baur2021autoencoders,zimmerer2019unsupervised} and diffusion models \citep{wyatt2022anoddpm} have been proposed.
Furthermore, anomaly detection has been successfully applied in chest X-ray images to detect COVID-19 \citep{zhang2020covid} or other malignancies \citep{lagogiannis_2023,mao2020abnormality}.

However, without exception, the existing works have followed the established paradigm of using the largest available training dataset, a notion we aim to challenge in this study.

\subsection{Core-set Selection}
To this end, we utilize methods from the field of core-set selection.
Core-set selection aims to create a small informative dataset such that the models trained on it show a similar test performance compared to those trained on the original dataset.
%
%
Core-set selection techniques for deep learning include minimizing the feature-space distance \citep{welling2009herding} or the distance of gradients with respect to a neural network \citep{mirzasoleiman2020coresets} between the selected subset and the original dataset.
In anomaly detection, core-set selection has been used by \citet{patchcore}.
Here, however, the selection is done on patch features instead of images.
In MemAE, \citet{gong2019memorizing} restricted the latent space of an autoencoder to a set of learned in-distribution feature vectors to perform anomaly detection.
While this work also finds prototypical feature vectors, they again cannot be linked back to training samples and, consequently, cannot be used for core-set selection.

\section{Surfacing Prototypical In-Distribution Samples Through Core-Set Selection}
\label{sec:method}

In unsupervised anomaly detection, the task is to train a model that can determine the binary label $y \in Y$ (ID: $y=-1$ or OOD: $y=1$) of a sample $x \in X$.
In this setting, it is commonly assumed that the training dataset $X_{\text{train}}$ contains only ID samples ($y = -1, \forall x \in X_{\text{train}}$), alleviating large labeling efforts for anomaly detection models.
Without loss of generality, a typical neural network used for anomaly detection can be divided into two parts:
The feature-extractor $\psi$ transforms a sample $x$ to its latent representation $z \in Z$, and a predictor $\phi$ that computes the anomaly score $s$ from $z$.
The anomaly score $s \in \mathbb{R}$ is a potentially unbounded continuous value that represents the \say{outlierness} of a sample $x$.
Combined, the anomaly detection model $\theta$ computes the anomaly score of a sample:
\begin{equation}
    \theta(x) = \phi(\psi(x)) = s \,.
\end{equation}

Such anomaly detection models are usually trained on large datasets.
When performing core-set selection, we want to determine a subset $X_{\text{sub}}$ with $M \in \mathbb{N}$ samples from the original training dataset $X_{\text{train}}$ with $N \in \mathbb{N}$ samples denoted as $x_i \in \mathbb{R}^D$ with $i = 1, 2, \ldots, N$, such that:

\begin{align}
    \begin{split}
        &\text{minimize} \quad E(X_{\text{sub}}, \theta) \quad \text{subject to} \\
        &\quad X_{\text{sub}} \subset X_{\text{train}}, \\ 
        &\quad |X_{\text{sub}}| = M,
    \end{split}
\end{align}

\noindent where $E(X_{\text{sub}}, \theta)$ is the detection error produced by a model $\theta$ trained on $X_{\text{sub}}$.

Previous work by \citet{padim} has shown that the latent space $Z_{\text{train}}$ represents a semantically meaningful compression of the training distribution $X_{\text{train}}$ that can be modeled as a multivariate Gaussian.
Due to the limited representational power of unimodal Gaussian distributions, we extend this idea by instead fitting a Gaussian Mixture Model (GMM) with $M$ components to this latent space.
Let $\mathcal{G} = \{ \mu_1, \mu_2, \ldots, \mu_M \}$ represent the set of means (centroids) of the $M$ components of the GMM, where $\mu_j$ denotes the mean of the $j$-th component.
This allows us to choose the samples corresponding to the latent codes that are closest to the centroids as core-set samples.

\begin{equation}
    X_{\text{sub}} = \{x_i \mid \forall \mu_j \in \mathcal{G}, \argmin_{x_i \in X_{\text{train}}} || \psi(x_i) - \mu_j ||_2 \} \,.
\end{equation}
The use of a GMM ensures that multiple different modes of normality are represented in $X_{\text{sub}}$.

\subsection{Weakly-Supervised Baselines}

In addition to the unsupervised core-set selection, we are interested in finding the optimal training subsets when labeled data is available.
Finding these subsets would offer additional insights into \say{normality} in anomaly detection and enable us to put the performance achieved by our core-set selection strategy into perspective, establishing an upper bound.
However, the problem is $\mathcal{NP}$-hard.
For a subset size of $M$, there exist $C(N, M) = \binom{N}{M} = \frac{N!}{M!(N-M)!}$ possibilities.
We therefore propose two approximate solutions to this problem, where we make use of the following simplification:
By evaluating the detection error obtained from training with individual samples, we aim to identify small yet high-performing training datasets.

\subsubsection{Greedy Selection}\label{sec:greedy} Since it is possible to train UAD models with only one sample, we can heuristically estimate the quality of each sample individually as $E(\{x_i\}, \theta, X_{\text{val}}) \text{ for } i = 1, 2, ..., N$.
From this information, we construct $X_{\text{sub}}$ as:

\begin{equation}
    X_{\text{sub}} = \argmin_{ \{ x_j | x_j \in X_{\text{train}} \text{, } 1 \leq j \leq M \} } \sum_{j=1}^M E(\{x_j\}, \theta, X_{\text{val}}) \,.
\end{equation}

In our work, we select the AUROC as the optimization target $E$.
Note that the set of samples that produce the smallest errors is not necessarily equal to the set of samples that together produce the lowest error.

\subsubsection{Evolutionary Algorithm} While the greedy approach above is intuitive, fast, and easy to implement, it prefers subsets of visually similar samples, as we will later show.
This is desirable in some cases; however, there are also scenarios in which multiple modes of normality should be covered by the selected subset.
To get a better coverage of the normal variations in a dataset, we propose a second approach, as described in the following.
For each combination of a training sample $x_i \in X_{\text{train}}$ and a validation sample $x_k \in X_{\text{val}}$, we compute an anomaly score $s(x_i,x_k)$ by training the anomaly detection model on $x_i$ only and running inference on $x_k$. The objective is to find a subset $X_{\text{sub}} = \{x_i | x_i \in X_{\text{train}}, 1 \leq i \leq M\}$ that maximizes a fitness function $f$ described as:

\begin{equation}\label{eq:fitness}
    f(X_{\text{sub}}) = \sum_{x_k, y_k \in X_{\text{val}}} \max_{x_i \in X_{\text{sub}}} y_k \cdot s(x_i, x_k) \,.
\end{equation}

Maximizing $f$ allows for finding $M$ training samples $x_i$ that achieve the best performance in classifying validation samples $x_k$ as ID or OOD. 
Since this problem is also $\mathcal{NP}$-hard, we approximate the solution using an evolutionary algorithm:

\begin{algorithm}
    \caption{Evolutionary Algorithm}
    \label{alg:evolutionary_algorithm}
    \KwData{Training dataset $X_{\text{train}}$, validation dataset $X_{\text{val}}$, population size $P$, anomaly scores $s(x_i,x_k) \, \forall x_i \in X_{\text{train}}, x_k \in X_{\text{val}}$, fitness function $f$, number of generations $G$}
    \KwResult{Approximately optimal subset $X^*_{\text{sub}}$}
    
    Initialize a random population $\mathcal{P} = \{X_{\text{sub}, p} | 1 \leq p \leq P\}$;
    
    \For{$gen \leftarrow 1$ \KwTo $G$}{
        Evaluate the fitness function $f$ for each individual $X_{\text{sub}, p} \in \mathcal{P}$\;
        Remove the least-fit $\frac{P}{2}$ individuals $X_{\text{sub}, p} \in \mathcal{P}$ from $\mathcal{P}$ to determine the best subset $\mathcal{P}'$\;
        Randomly apply either a crossover (combine two individuals) or mutation (replace one sample) operation to each individual in $\mathcal{P}'$ to create a modified population $\mathcal{P}''$\;
        Generate a new population $\mathcal{P} = \mathcal{P}' \cup \mathcal{P}''$\;
    }
    \Return{Best individual found in the final population}\;
\end{algorithm}

In the crossover operation, random subsets of two individuals $X_{\text{sub}, 1}$, $X_{\text{sub}, 2} \in \mathcal{P}'$ (called parents) are merged to produce a new individual $X'_{\text{sub}}$, such that $|X'_{\text{sub}}| = M$.
In the mutation operation, one sample $x_1 \in X_{\text{sub}}$ of an individual $X_{\text{sub}} \in \mathcal{P}'$ is randomly replaced with another sample $x_2 \in D \setminus X_{\text{sub}}$ to produce a new individual $X'_{\text{sub}}$:

\begin{equation}
    X'_{\text{sub}} = X_{\text{sub}} \setminus \{ x_1 \} \cup \{ x_2 \} \,.
\end{equation}

In contrast to the greedy selection strategy that favors visually similar samples, the subsets found by the evolutionary algorithm have better coverage of the different notions of normality contained in the training dataset (c.f. \Cref{fig:greedy_evo_screw}).
However, note that this enhanced coverage could also be harmful when the normal dataset is noisy and contains samples that should be considered abnormal.
In such cases, greedy selection is more effective at filtering out these samples.
\section{Experiments}
\label{sec:experiments}

\subsection{Datasets and Models}\label{sec:datasets_and_models}

To evaluate our methods, we use datasets from the natural- and medical-image domains, showing the applicability of our method in diverse tasks.
CIFAR10, CIFAR100 \citep{cifar10}, MNIST \citep{mnist}, and Fashion-MNIST \citep{fashion_mnist} are trained in a one-vs-rest setting, where one class is used as the in-distribution, and all other classes are combined as outliers.
MVTec-AD \citep{mvtecad} is an industrial defect detection dataset and a frequently used benchmark for UAD models.
In the chest X-ray images of the RSNA Pneumonia Detection dataset \citep{rsna}, the in-distribution constitutes images of healthy patients, while anomalous samples show signs of pneumonia or other lung opacities.
In addition, we use CheXpert \citep{chexpert} to test if the samples found by our method generalize to other datasets.
Similarly to RSNA, the in-distribution samples here are images labeled with \say{No Finding}, while OOD samples either display pneumonia or other lung opacities.
Lastly, we detect MRI slices with glioma in the BraTS dataset \citep{brats1,brats2}.
For each dataset, we chose a respective state-of-the-art model: PANDA \citep{panda} is used for CIFAR10, CIFAR100, MNIST, and Fashion-MNIST, PatchCore \citep{patchcore} for MVTec-AD, and FAE \citep{fae,lagogiannis_2023} for RSNA, CheXpert and BraTS.
We additionally used Reverse Distillation (RD) by \citet{rd} for RSNA to test the generalizability of identified samples across models.
Details about the datasets and models can be found in the supplementary material.

\subsection{Experimental Setup}\label{sec:experimental_setup}

We carefully tune all baselines and methods using established protocols where applicable.
Details are in \Cref{sec:implementation_details}.
We restricted the maximum subset size to $M = 25$ samples in our experiments as we did not experience substantial increases in performance beyond this point. Multiplied by the number of experiments and selection strategies, this decision saved significant amounts of our limited resources and allowed for more extensive experimentation in other dimensions.

\section{Results and Discussion}\label{ref:discussion}

In the following, we first present our main findings in \Cref{sec:few_samples}, \Cref{sec:what_is_normal}, and \Cref{sec:transfer_models_datasets}.
Then, we provide theoretical justification for these findings (\Cref{sec:long_tail_discussion,sec:should_outliers}) and dive deeper into the specific characteristics of the different datasets and how they impact the core-set selection performance in \Cref{sec:differences_in_normal_and_abnormal}.

\subsection{A Few Selected Samples Can Outperform Training With the Whole Dataset}\label{sec:few_samples}

\begin{table*}[t]
    \centering
    \small
    \caption{AUROC scores of training with the full dataset and training with 1, 5, 10, and 25 best-performing (with greedy search, the evolutionary algorithm, and core-set selection) or random samples. Since the performance of randomly selected subgroups can vary strongly, we repeated these experiments over ten different subsets. Bold numbers show the best performance per dataset, and underlined numbers are the best per sample size. Numbers marked with * surpass training with $100\%$ of the data. This table shows the averaged results over each class in the respective datasets. Detailed results for all classes can be found in the Appendix.}
    \resizebox{\linewidth}{!}{
    \begin{tabular}{@{}lcccccccc@{}}
        \\
        \\
        \toprule 
        & \thead{Method} & \thead{CIFAR10} & \thead{CIFAR100} & \thead{F-MNIST} & \thead{MNIST} & \thead{MVTec-AD} & \thead{BraTS} & \thead{RSNA} \\ 
        \midrule 
        \multirow{4}{*}{\begin{sideways}1 sample\end{sideways}}
        & Random & 84.96 \std{1.2} & 77.05 \std{1.4} & 83.51 \std{3.3} & 76.11 \std{2.2} & 83.61 \std{1.2} & 93.85 \std{5.3} & 61.86 \std{5.2} \\
        & Greedy & \underline{95.02} & \underline{89.94} & \underline{94.00} & \underline{90.15} & \underline{89.70} & 95.25 & \underline{70.64} \\ 
        & Evo & 87.68 & 79.99 & 86.82 & 74.82 & 85.55 & 94.00 & 65.62 \\
        & Core-set & 90.08 & 79.62 & 91.47 & 82.29 & 82.80 & \underline{96.25} & 67.36 \\
        \midrule
        \multirow{4}{*}{\begin{sideways}5 samples\end{sideways}}
        & Random & 91.95 \std{0.7} & 86.40 \std{0.8} & 91.86 \std{1.1} & 89.36 \std{1.1} & 90.43 \std{0.9} & 97.83 \std{1.1} & 73.14 \std{3.5} \\
        & Greedy & \underline{96.36} & \underline{92.49} & \underline{95.31} & \underline{93.62} & 91.52 & 97.25 & 74.19 \\ 
        & Evo & 94.30 & 90.40 & 93.42 & 92.96 & \underline{94.12} & \underline{99.06}* & \underline{76.45} \\
        & Core-set & 93.85 & 89.74 & 93.89 & 92.55 & 92.13 & 96.88 & 75.06 \\
        \midrule
        \multirow{4}{*}{\begin{sideways}10 samples\end{sideways}}
        & Random & 93.59 \std{0.4} & 89.05 \std{0.5} & 93.24 \std{0.3} & 93.00 \std{0.6} & 92.80 \std{0.6} & 97.96 \std{1.0} & 75.43 \std{3.0} \\
        & Greedy & \underline{96.37} & \underline{92.59} & \underline{95.38} & 94.91 & 92.94 & 97.06 & \underline{77.80} \\ 
        & Evo & 95.51 & 91.78 & 94.28 & \underline{95.37} & 96.37 & \underline{98.81}* & 76.61 \\
        & Core-set & 94.28 & 91.03 & 94.47 & 94.94 & \underline{96.77} & 98.19 & 76.78 \\
        \midrule
        \multirow{4}{*}{\begin{sideways}25 samples\end{sideways}}
        & Random & 94.68 \std{0.2} & 91.52 \std{0.2} & 94.16 \std{0.2} & 96.24 \std{0.3} & 95.29 \std{0.3} & 98.09 \std{0.6} & 77.08 \std{0.9} \\
        & Greedy & \underline{96.29} & 92.60 & \underline{95.52} & 96.06 & 93.53 & 97.94 & 78.60* \\
        & Evo & 95.51 & 91.88 & 95.07 & 97.30 & \textbf{\underline{98.52}}* & 98.87* & \textbf{\underline{80.59}}* \\
        & Core-set & 94.85 & \underline{92.88} & 95.13 & \underline{97.33} & 98.37 & \textbf{\underline{99.12}}* & 79.18* \\
        \midrule 
        & Full training & \textbf{96.58} & \textbf{94.92} & \textbf{95.79} & \textbf{98.41} & 98.48 & 98.75 & 77.97 \\
        \bottomrule
    \end{tabular}}
    \vspace{2pt}
    \label{tab:results_all}
\end{table*}

As shown in \Cref{tab:results_all}, UAD models achieve high performance even when trained with only a few samples.

\begin{wrapfigure}{l}{0.5\textwidth}
    \centering
    \vspace{-5pt}
    \includegraphics[width=\linewidth]{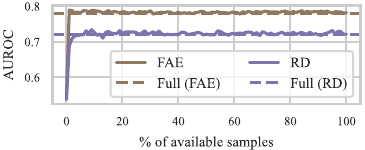}
    \caption{Performance in RSNA does not increase with more training samples.}
    \vspace{-5pt}
    \label{fig:rsna_all}
\end{wrapfigure}
\noindent Surprisingly, even randomly selected subsets can result in a strong model on all considered datasets, confirming our hypothesis that UAD models do not overfit.
\Cref{fig:rsna_all} depicts a typical observation that detection performance saturates already with very few samples, regardless of the UAD model used.
However, not all random subsets perform equally well, which can be seen by the large standard deviations (up to $15.1$) for many categories (see \Cref{tab:results_cifar10} in \Cref{app:detailed_results}).
Our proposed core-set selection strategy substantially outperforms random selection on all datasets and even performs better than full training on BraTS and RSNA.
Notably, for the latter, it uses as little as $0.3\%$ of the available training data.
In addition to this, the evolutionary algorithm and especially the greedy selection strategy also demonstrate good performance.
Moreover, our core-set selection strategy is not far behind and outperforms the other two with 25 samples on CIFAR100 and MNIST despite not using any labels.
Overall, our selection strategies outperform full training in $25/67$ categories tested in this study (see \Cref{tab:results_cifar10,tab:results_cifar100,tab:results_fashion_mnist,tab:results_mnist,tab:results_mvtec} in \Cref{app:detailed_results}).
Furthermore, the gap between random and informed selection becomes even more pronounced in the very-low data regime (1--10 samples).


\subsection{What Characterizes Normal Samples?}\label{sec:what_is_normal}

\begin{figure}[htb]
    \centering
    \includegraphics[width=\textwidth]{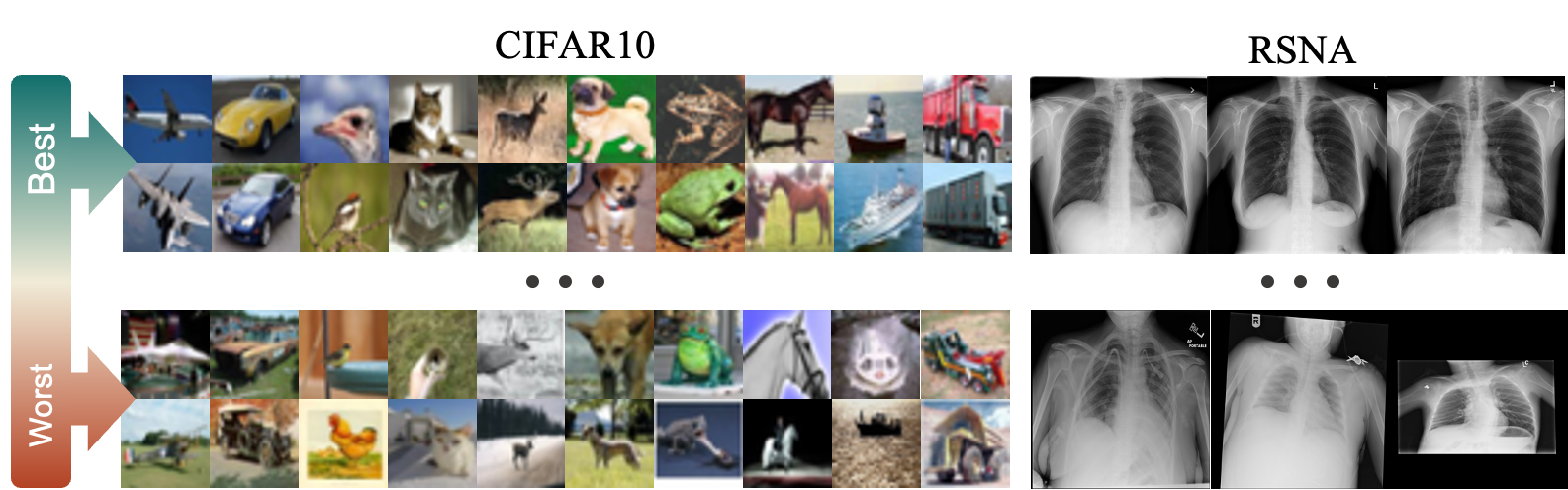}
    \caption{Best- and worst-performing samples in CIFAR10 and RSNA. Identified using our proposed core-set selection strategy.}
    \label{fig:best_worst}
\end{figure}

Our method not only allows training strong UAD models with only a few samples, but it also provides insights into what constitutes a prototypical in-distribution image.
\Cref{fig:best_worst} shows the best- and worst-performing samples for each class in CIFAR10 on the left.
The \say{best} images display well-lit prototypical objects that are well-centered, have good contrast, and have mostly uniform backgrounds.
In contrast, the \say{worst} in-distribution images include drawings (bird and horse), toys (frog, truck), historical objects (plane, car), or images with bad contrast (dog).
In noisy, less well-curated datasets like RSNA, our method can effectively detect and filter low-quality samples.
\Cref{fig:best_worst} reveals severe deformations, foreign objects such as access tubes or implants, low tissue contrast, or dislocations in the worst-performing samples.
The best-performing ones, on the other side, are well-centered and detailed, contain male and female samples, and clearly show the lungs, a prerequisite for detecting pneumonia.

Motivated by the insights we gained about the characteristics of in-distribution samples, we manually selected a training subset.
We chose the RSNA dataset for this experiment because, unlike MVTec-AD, it contains atypical samples in the \say{normal} data and because the images are large enough to be visually inspected, in contrast to the other natural image datasets.
We selected samples that had similar characteristics as displayed in \Cref{fig:best_worst} and covered the distribution of ID samples well.
We only started the evaluation of the manually selected samples once their selection was complete.
No information other than the characteristics described above was used for the manual selection, and the author selecting the samples is not a trained radiologist or other medical expert.
When training with these manually selected samples, we achieved AUROCs of $\mathbf{67.88}$, $\mathbf{76.61}$, $\mathbf{79.73}$, and $\mathbf{81.04}$ for 1, 5, 10, and 25 samples, respectively.
The manual selection strategy, therefore, outperformed all automatic selection strategies and even full training, giving the best results on RSNA in this work.

We repeated a similar experiment for the -- arbitrarily selected -- \say{8} class on MNIST.
This time, however, we did not look at the best- or worst-performing samples from this class but simply selected samples that are visually close to a prototypical 8.
We discarded samples that had non-closed lines, where the curves of the 8 were excessively slim, and those with irregular or wavy lines.
With these samples, we achieved AUROCs of $\mathbf{76.69}$, $\mathbf{92.42}$, $\mathbf{94.59}$, and $\mathbf{96.37}$ for 1, 5, 10, and 25 samples, respectively, outperforming both random selection and the evolutionary strategy, while also only falling $1.45$ points below full training performance.

\subsection{Prototypical Samples Are Transferrable to Other Models and Datasets} \label{sec:transfer_models_datasets}

\begin{figure}[ht]
    \centering
    \includegraphics[width=\textwidth]{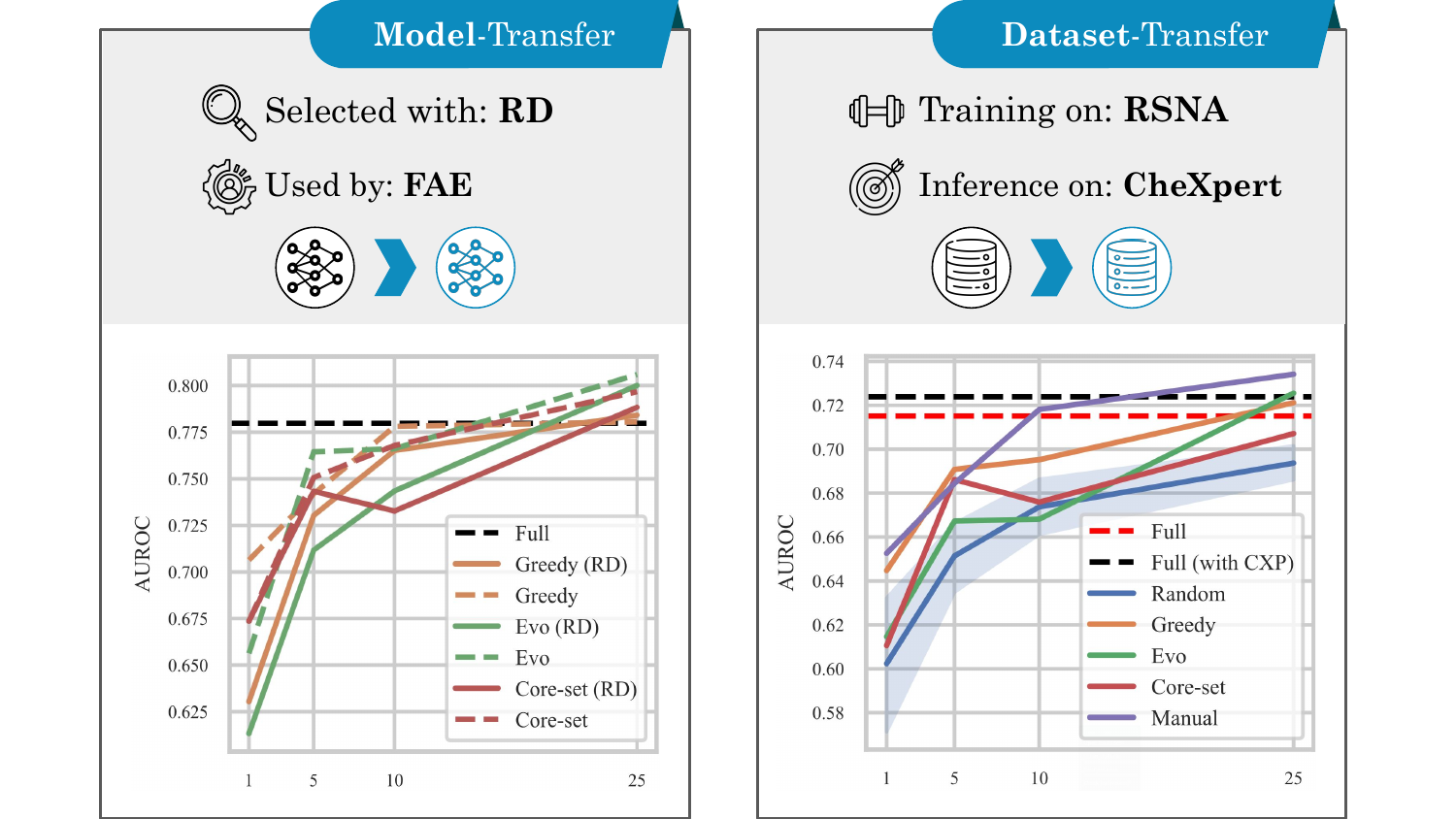}
    \caption{Prototypical samples transfer well to other datasets and models. \textbf{Left}: Surfaced samples with the RD model achieve high performance when used with FAE. Test performance of FAE on RSNA when samples are selected using RD (full lines) or FAE (dashed lines). \textbf{Right}: Training with 25 carefully selected samples from RSNA can exceed full training performance with CheXpert (8443 samples) when evaluated on the latter. Test performance on CheXpert after training on CheXpert samples (black, dashed line) or RSNA (other lines).}
    \label{fig:transfer}
\end{figure}
We also trained a second type of UAD model, Reverse Distillation (RD), on RSNA.
This model matches encoder and decoder representations at different levels.
The left side of \Cref{fig:transfer} shows how the best-performing samples selected with our proposed core-set selection and the two weakly-supervised baselines on RD perform when applied to the FAE model. 
Although RD generally performs worse than FAE, its best-performing set of samples works well on FAE, performing almost on par with the ones found with FAE itself and even exceeding full performance.
We conclude from this result that there are commonalities between the best samples that are independent of the model.
This means that samples found using one model can be transferred to another.

Similarly, high performance for sample combinations on RSNA translates well to the CheXpert dataset.
As expected, training with RSNA samples gives a slightly lower performance on CheXpert than training on CheXpert itself (red and black dashed lines in \Cref{fig:transfer}, right).
This gap, however, can be closed by training with only 25 high-performing samples from RSNA.
Even more impressive, we reached higher performance on CheXpert when training with the 25 manually selected RSNA samples than when training on the full CheXpert dataset itself.



\subsection{Long-Tail In-Distribution Samples Can Degrade Anomaly Detection Performance} \label{sec:long_tail_discussion}

\begin{figure*}[ht]
    \centering
    \includegraphics[width=.9\linewidth]{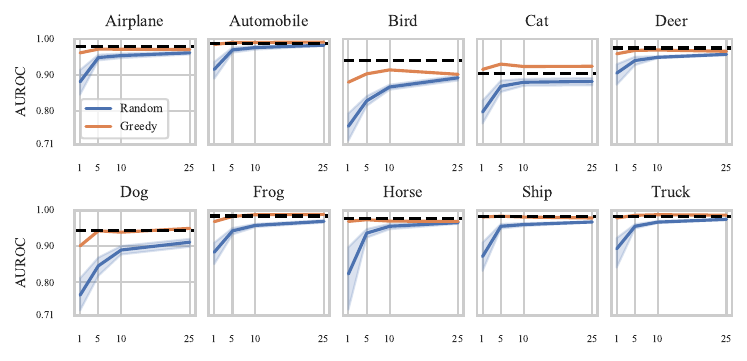}
    \caption{Only 10 representative training samples are needed to surpass the performance of training with the whole dataset on five out of the ten classes in CIFAR10. AUROC for training with 1, 5, 10, and 25 random or best (greedy selection) samples on CIFAR10. For random samples, the experiments were repeated ten times with different samples. The dashed black line represents training with all 4000 ID samples.}
    \label{fig:results_cifar10}
\end{figure*}

\begin{figure}[htb]
    \centering
    \begin{minipage}[t]{.475\textwidth}
        \includegraphics[width=\linewidth]{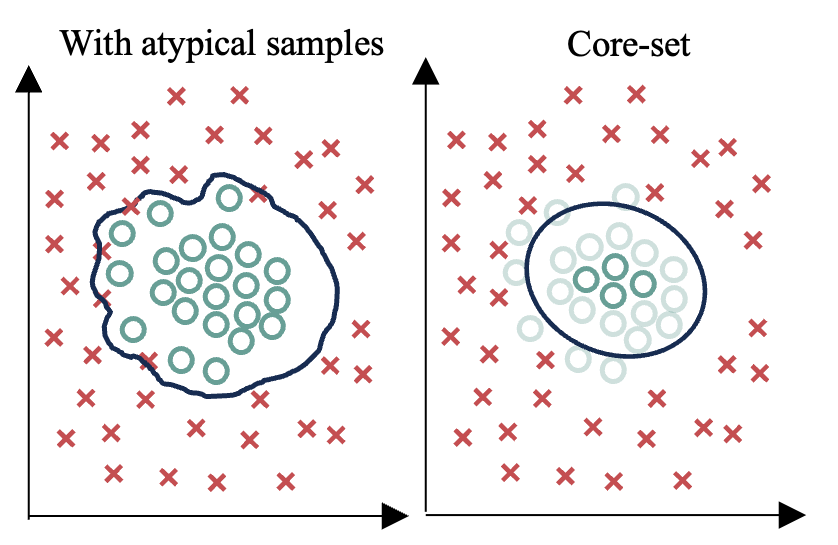}
        \caption{Training with only a few selected samples helps to ignore training samples that lie at the tails of the in-distribution. Illustration of our hypothesis of how long-tail in-distribution samples can skew the decision boundary. \textbf{Left}: Some ID samples might be closer to the OOD data and shift the decision boundary. \textbf{Right}: Training with a few carefully selected samples creates a better decision boundary.}
        \label{fig:small_sample_hypothesis}
    \end{minipage}
    \hfill
    \begin{minipage}[t]{.475\textwidth}
        \includegraphics[width=\linewidth]{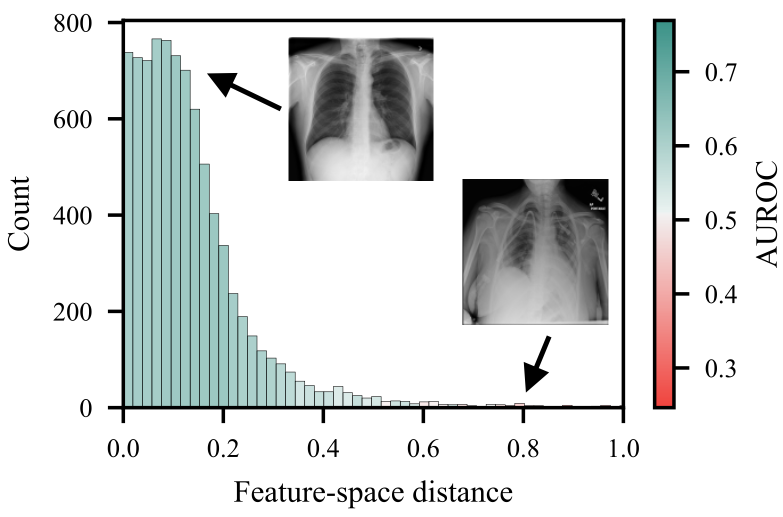}
        \caption{RSNA follows a long-tail distribution in the FAE feature-space, and the samples at the tails perform worse. Histogram of distances of all training samples from RSNA to the center of an FAE model and two example images from the center and the tail. The bins of the histogram are colored-coded according to the AUROCs of each single sample as described in \Cref{sec:greedy}.}
        \label{fig:panda_distances}
    \end{minipage}
\end{figure}

In \Cref{tab:results_all,fig:rsna_all,fig:transfer}, we have seen that very small datasets can exceed the performance of full training.
\Cref{fig:results_cifar10} even shows that peak performance for the \say{cat} class of CIFAR10 is achieved with only five samples.
These results suggest that there exist samples in many datasets whose inclusion degrades performance.
We hypothesize that the reason for this is due to the nature of the in-distributions, which often have long tails, as shown by \citet{feldman2020does} and \citet{zhu2014capturing}.
The long-tail hypothesis states that the majority of the in-distribution samples have only low inter-sample variance, with the exception of a few rare samples that differ a lot from the rest (while still being part of the in-distribution).
While these samples can be actively contrasted to other classes and memorized by supervised machine learning models \citep{feldman2020neural}, such mechanisms are not available for UAD models, where the long-tail in-distribution samples are treated as any other training sample.
This can shift the decision boundary in an unfortunate way (\Cref{fig:small_sample_hypothesis}, left).
Training with carefully selected samples effectively ignores these data points and can lead to better performance despite using fewer samples (\Cref{fig:small_sample_hypothesis}, right).
\Cref{fig:panda_distances} reveals that the datasets used in our study also follow a long-tail distribution and that the samples at the tails perform worse.
Additionally, the worst-performing samples in \Cref{fig:best_worst} are clearly atypical and, thus, likely also lay at the tail of the in-distribution.

\subsection{Should Samples From the Long Tails of the In-Distribution Be Considered Outliers?} \label{sec:should_outliers}

Our experiments suggest that there are samples in the training datasets that lie at the tails of the in-distribution and lower the performance of the UAD model.
Our subset-selection strategies are effective at filtering out these data points.
Of course, ignoring such long-tail samples during training will declare them as outliers, which, at first sight, is false given the labels.
We argue, however, that these data points should be considered as such because they warrant special consideration in downstream tasks. For example, a subsequent supervised classification algorithm is more likely to misclassify these, and in such a scenario, flagging long tail samples as OOD is desired.
Atypical X-ray images, as shown on the bottom right of \Cref{fig:best_worst}, can pose difficulties (even for manual diagnosis) and should also receive special attention.
Further, as we have shown, including these samples during training can lower the classification performance for other samples that are then falsely flagged as ID.
The selection methods presented in this study can, therefore, not only be used to identify the most prototypical in-distribution samples but can also be used to automatically filter a dataset from noisy or corrupted images.

\subsection{The Distributions of Normality and Abnormality Differ Between Datasets} \label{sec:differences_in_normal_and_abnormal}

\begin{figure}[ht]
    \centering
    \includegraphics[width=\linewidth]{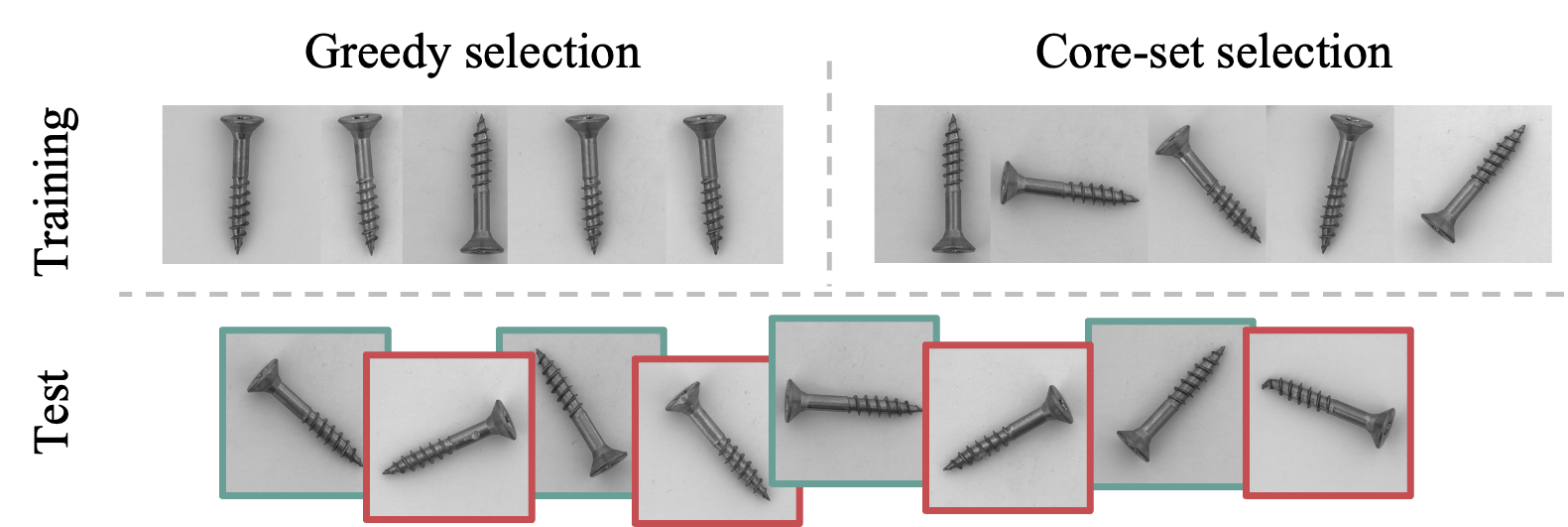}
    \caption{Greedy selection favors visually similar samples, while the core-set selection achieves a better coverage of the normal variations in the training dataset. Best five training samples for the \textit{screw} category in MVTec-AD, found using greedy selection (left) and the unsupervised core-set selection (right). Bottom: normal (green) and defective (red) samples in the test set.}
    \label{fig:greedy_evo_screw}
\end{figure}

Despite all being used for anomaly detection and showing similar behavior regarding training with few samples, the datasets considered in our study vary greatly with respect to their in and out distribution, as well as the relationships between the two.
While for some datasets, the variability between ID samples is comparably low; other datasets contain multiple modes of normality.
The objects within each class of the MVTec-AD dataset are very similar and often only oriented differently, and the chest X-ray images all show the same anatomical region.
The brain MR images in BraTS are registered to an atlas and, consequently, have even lower anatomical variance.
This is in contrast to CIFAR10, CIFAR100, and Fashion-MNIST, where the in-distribution training class can exhibit various shapes, poses, and colors.
Similarly, OOD samples can be local and subtle, as in MVTec-AD, BraTS, and RSNA, or global, as in CIFAR10, CIFAR100, MNIST, Fashion-MNIST.
A good subset of prototypical samples should cover the different modes of normality but exclude the samples that are atypical and semantically too close to the out-distribution.
When looking at examples of the different types of in- and out-distributions described above, we can identify the strengths and weaknesses of the different selection strategies.
An example of a dataset with different ID modes and subtle anomalies is the \say{screw} category of MVTec-AD.
\Cref{fig:greedy_evo_screw} shows that greedy selection favors combinations of similarly oriented images and fails to cover the whole space of differently-oriented, normal samples.
Our proposed unsupervised core-set selection strategy, on the other hand, (and also the evolutionary algorithm) better covers the different orientations that should all be considered normal (see also \Cref{tab:results_mvtec} in \Cref{app:detailed_results}).

\section{Conclusion}\label{ref:conclusion}

In many domains of Deep Learning, the prevailing assumption is that more training data leads to better models.
Our work challenges this practice for UAD and highlights the importance of data quality over data quantity.
Specifically, we have shown that it is a common phenomenon of UAD models to achieve state-of-the-art performance with extremely few training samples.
Our proposed core-set selection strategy is a fast and easy-to-implement, unsupervised approach to automatically extract such high-performing, prototypical training samples.
We have further shown that these prototypical in-distribution samples are transferrable across models, datasets, and even modalities and tasks and that the characteristics derived from them allow for manual selection.
This understanding of the well-performing training samples can help in designing better UAD models in the future.
Our findings further make UAD models much more attractive in practice despite their performance still lacking behind that of their supervised counterparts.

\bibliography{main}
\bibliographystyle{tmlr}

\clearpage
\appendix

\section{Appendix: Datasets}
\label{sec:suppl_datasets}

\subsubsection{CIFAR10, CIFAR100, MNIST, and FashionMNIST}
We follow the widely used training setup from \citet{panda} for these datasets.
CIFAR10 contains $6000$ images per class.
For each class $c \in C$, we create a training dataset $X_\text{train, c}$ using $4000$ samples from said class.
The remaining samples are split equally into a validation and a test set and combined with the same amount of samples from every other class as outliers.
Training, validation, and test sets for CIFAR100, MNIST, and Fashion-MNIST are created analogously.
For CIFAR100, we used the 20 superclasses instead of the 100 detailed classes.

\subsubsection{MVTec-AD}
MVTec-AD is a dataset for defect detection in industrial production. Here, we used the original splits as outlined by \citet{mvtecad}.

\subsubsection{BraTS}
The multimodal brain tumor image segmentation benchmark (BraTS) contains 369 MRI from patients with glioma.
We extract 5 slices around the center line and use 101 slices without glioma for training.
The remaining 80 normal slices are split 50:50 into a validation and a test set and are complemented with 40 pathological slices each.
Following \citet{fae}, we use the T2-weighted sequences, perform histogram equalization on the slices, and resize them to $128 \times 128$.

\subsubsection{RSNA and CheXpert}
The RSNA Pneumonia Detection dataset \citep{rsna} is a subset of $30\,000$ frontal view chest radiographs from the National Institutes of Health (NIH) CXR8 dataset that was manually labeled by 18 radiologists for one of the following labels: \say{Normal}, \say{Lung Opacity}, or \say{No Lung Opacity / Not Normal}.
The CheXpert database contains $224\,316$ chest-radiographs of $65\,240$ patients, acquired at Stanford Hospital with 13 structured diagnostic labels.
To make the CheXpert compatible with RSNA, we only considered frontal-view images without support devices and further excluded those where any of the labels were marked as uncertain.
In addition to the image data and labels, demographic information about the patients' gender and age was available for both datasets.

For RSNA, we used the \say{Normal} label as in-distribution images and combined the \say{pneumonia} and \say{No Opacity/Not Normal} (has lung opacities, but not suspicious for pneumonia) as OOD.
Similarly, for CheXpert, the \say{No Finding} label was used as in-distribution, and \say{pneumonia} and \say{lung opacity} as OOD.
For both datasets, we created a validation- and a test set that are balanced w.r.t. gender (male and female), age (young and old), the presence of anomalies, and contain 800 samples each.
The remaining in-distribution samples were used for training.
As part of preprocessing, all Chest X-ray images were center cropped and resized to $128 \times 128$ pixels.
Note that we treated both datasets individually and did not combine them for training or evaluation.

\section{Appendix: Models}
\label{sec:suppl_models}

\subsubsection{PANDA} PANDA \citep{panda} is a model built upon DeepSVDD by \citet{deepsvdd}.
Like the latter, it relies on training a one-class classifier by using the compactness loss.
Given a feature-extractor $\psi$ and a center vector $c$, the compactness loss is defined as:

\begin{equation}
    \mathcal{L}_{\text{compact}} = \sum_{x \in D} || \psi(x) - c ||^2 \,.
\end{equation}

The center vector $c$ is computed as the mean feature vector of the training dataset $D$ on the untrained model $\psi_0$:

\begin{equation}
    c = \frac{1}{|D|} \sum_{x \in D} \psi_0(x) \,.
\end{equation}

Instead of training a specialized architecture from scratch, PANDA benefits from useful features of pre-trained models.
Specifically, it extracts features from the penultimate layer of a ResNet152 \citep{resnet} and categorizes samples into ID / OOD by the L1-distance to the $k$ nearest neighbors (kNN), with $k = 2$.
PANDA only fine-tunes \texttt{layer3} and \texttt{layer4} and uses early stopping to determine the optimal distance between ID and OOD samples.
Training has no effect when using only one sample as the feature representation of the only sample is always identical to $c$.
We used the original configuration from their paper for our experiments.

\subsubsection{PatchCore} PatchCore \citep{patchcore} follows a similar concept as PANDA.
However, instead of performing a kNN search on pooled, global features, PatchCore achieves localized anomaly detection by using a memory bank of locally-aware patch features $\mathcal{M}$ instead.
To make the kNN search computationally feasible, PatchCore performs core-set selection on the memory bank to retain only a subset of representative features $\mathcal{M}_\text{sub}$:

\begin{equation}
    \mathcal{M}^*_\text{sub} = \argmin_{\mathcal{M}_\text{sub} \subset \mathcal{M}} \max_{m \in \mathcal{M}} \min_{n \in \mathcal{M}_\text{sub}} || m - n ||_2 \,.
\end{equation}

Compared to PANDA, PatchCore does not require fine-tuning of the feature extractor.
We used the same hyperparameters for PatchCore as in the original publication.

\subsubsection{FAE} The Structural Feature-Autoencoder (FAE) \citep{fae} extracts spatial feature maps from a pre-trained and frozen feature extractor $\psi$ (a ResNet18 \citet{resnet} in practice).
The feature maps a resized and concatenated and fed into a convolutional autoencoder $f_{\theta}$ that is trained via the structural similarity (SSIM) loss for reconstruction:

\begin{equation}
    \mathcal{L} = \text{SSIM}(\psi(x), f_\theta(\psi(x))) \,.
\end{equation}

Anomalies are detected using the residual between the feature maps and their reconstruction like in popular image-reconstruction models.
It was found to perform best for UAD in Chest X-ray images in a study by \citet{lagogiannis_2023}.
We use a smaller model in our experiments since it still gave us the same performance on the RSNA full dataset as the larger model and is more resource-efficient.
Specifically, we set the \verb|fae_hidden_dims| parameter to \verb|[50, 100]|.
The other parameters were kept the same as in the original publication.

\subsubsection{Reverse Distillation} The Reverse Distillation (RD) model by \citet{rd} utilizes a frozen encoder model and a decoder that mirrors the former, similar to an Autoencoder.
Instead of reconstructing the image, however, RD minimizes the cosine distance between feature maps in the encoder and decoder.
The same measure is also used during inference to detect anomalies.
RD was the second-best performing UAD model for Chest X-ray images in \citet{lagogiannis_2023}.
We used the same hyperparameters for RD as in \citet{lagogiannis_2023}.

\section{Appendix: Implementation details}\label{sec:implementation_details}

As suggested by \citet{panda}, we train PANDA for a constant number of $2355$ steps (corresponding to 15 epochs on CIFAR10) using the SGD optimizer with a learning rate of $0.01$ and weight decay of $0.00005$.
FAE converges very fast on BraTS, RSNA, and CheXpert, so it was trained for $500$ steps using the Adam optimizer with a learning rate of $0.0002$.
In our experiments, the evolutionary algorithm was robust to the population size $P$ and the number of generations $G$.
In all experiments, we therefore empirically set $P=1000$ and $G=500$.
For all models, we used the official PyTorch implementations by the authors.



\section{Appendix: Detailed results}\label{app:detailed_results}

We show the detailed per-class results for CIFAR10, CIFAR100, MNIST, Fashion-MNIST, and MVTec-AD in \Cref{tab:results_cifar10,tab:results_mnist,tab:results_fashion_mnist,tab:results_mvtec}, respectively.

\begin{table*}
    \centering
    \small
    \caption{Detailed training results for CIFAR10. AUROC scores of full training and training with 1, 5, 10, and 25 best-performing (with greedy search, the evolutionary algorithm, and core-set selection) or random samples. Since the performance of randomly selected subgroups can vary strongly, we repeated these experiments over ten different subsets. Best performances are marked in bold, and underlined numbers are the best per sample size.}
    \resizebox{\linewidth}{!}{
    \begin{tabular}{@{}lcccccccccccc@{}}
        \toprule 
        & \textbf{Method} & \textbf{Airplane} & \textbf{Automobile} & \textbf{Bird} & \textbf{Cat} & \textbf{Deer} & \textbf{Dog} & \textbf{Frog} & \textbf{Horse} & \textbf{Ship} & \textbf{Truck} & \textbf{Average} \\ 
        \midrule
        \multirow{4}{*}{\begin{sideways}1 sample\end{sideways}}
        & Random & 88.14 \std{6.3} & 91.59 \std{4.1} & 75.81 \std{6.2} & 79.75 \std{5.5} & 90.55 \std{4.7} & 76.47 \std{7.6} & 88.38 \std{5.1} & 82.34 \std{15.1} & 87.19 \std{6.6} & 89.32 \std{7.6} & 84.96 \std{1.2} \\ 
        & Greedy & \underline{96.18} & \underline{98.50} & \underline{88.04} & \underline{91.59} & \underline{95.92} & \underline{90.13} & \underline{96.87} & \underline{96.94} & \underline{98.14} & \underline{97.88} & \underline{95.02} \\ 
        & Evo & 86.15 & 97.08 & \underline{88.04} & 78.61 & 93.27 & 79.83 & 67.29 & 94.21 & 96.64 & 95.73 & 87.68 \\ 
        & Core-set & 94.54 & 96.86 & 79.14 & 81.56 & 92.75 & 84.63 & 92.84 & 92.28 & 92.09 & 94.15 & 90.08 \\ 
        \midrule 
        \multirow{4}{*}{\begin{sideways}5 samples\end{sideways}}
        & Random & 94.88 \std{1.3} & 97.00 \std{1.2} & 82.88 \std{2.4} & 86.92 \std{2.7} & 94.06 \std{2.0} & 84.61 \std{4.5} & 94.30 \std{1.5} & 93.68 \std{2.0} & 95.57 \std{1.1} & 95.58 \std{1.0} & 91.95 \std{0.7} \\ 
        & Greedy & \underline{97.22} & \underline{99.05} & \underline{90.36} & \textbf{\underline{93.06}} & \underline{96.86} & \underline{94.32} & \underline{98.31} & \underline{97.39} & \textbf{\underline{98.37}} & \underline{98.61} & \underline{96.36} \\ 
        & Evo & 95.85 & 96.60 & 89.19 & 87.44 & 96.12 & 92.23 & 93.89 & 97.11 & 96.72 & 97.88 & 94.30 \\ 
        & Core-set & 94.14 & 97.23 & 88.70 & 87.88 & 95.91 & 90.36 & 95.24 & 96.39 & 96.22 & 96.46 & 93.85 \\ 
        \midrule 
        \multirow{4}{*}{\begin{sideways}10 samples\end{sideways}}
        & Random & 95.42 \std{1.2} & 97.65 \std{0.8} & 86.70 \std{1.2} & 88.02 \std{1.6} & 94.94 \std{0.5} & 89.00 \std{2.0} & 95.82 \std{0.8} & 95.57 \std{1.3} & 96.05 \std{0.7} & 96.74 \std{0.5} & 93.59 \std{0.4} \\ 
        & Greedy & \underline{97.19} & \underline{99.08} & \underline{91.45} & \underline{92.39} & \underline{96.97} & \underline{93.94} & \underline{98.79} & \underline{97.01} & \underline{98.14} & \textbf{\underline{98.79}} & \underline{96.37} \\ 
        & Evo & 96.47 & 97.98 & 90.64 & 89.32 & 96.62 & 93.61 & 97.72 & 96.93 & 97.62 & 98.23 & 95.51 \\ 
        & Core-set & 94.90 & 97.79 & 88.56 & 88.09 & 95.91 & 91.50 & 96.31 & 96.46 & 96.45 & 96.81 & 94.28 \\ 
        \midrule 
        \multirow{4}{*}{\begin{sideways}25 samples\end{sideways}}
        & Random & 96.19 \std{0.7} & 98.30 \std{0.3} & 89.26 \std{1.2} & 88.26 \std{1.7} & 95.77 \std{0.5} & 91.16 \std{1.9} & 96.95 \std{0.6} & 96.59 \std{0.5} & 96.77 \std{0.2} & 97.50 \std{0.3} & 94.68 \std{0.2} \\ 
        & Greedy & \underline{97.11} & \textbf{\underline{99.17}} & 90.22 & \underline{92.46} & 96.61 & \textbf{\underline{95.00}} & \textbf{\underline{98.87}} & 96.93 & \underline{97.87} & \underline{98.65} & \underline{96.29} \\ 
        & Evo & 96.56 & 98.18 & \underline{91.38} & 89.16 & \underline{96.68} & 92.11 & 97.87 & \underline{97.42} & 97.64 & 98.08 & 95.51 \\ 
        & Core-set & 96.88 & 98.67 & 89.06 & 86.26 & 95.98 & 92.98 & 97.40 & 96.99 & 96.68 & 97.56 & 94.85 \\ 
        \midrule 
        & Full training & \textbf{97.92} & 98.76 & \textbf{94.12} & 90.43 & \textbf{97.53} & 94.36 & 98.44 & \textbf{97.83} & 98.21 & 98.23 & \textbf{96.58} \\ 
        \bottomrule
    \end{tabular}}
    \vspace{2pt}
    \label{tab:results_cifar10}
\end{table*}

\begin{table*}
    \centering
    \small
    \caption{Detailed training results for MNIST. AUROC scores of full training and training with 1, 5, 10, and 25 best-performing (with greedy search, the evolutionary algorithm, and core-set selection) or random samples. Since the performance of randomly selected subgroups can vary strongly, we repeated these experiments over ten different subsets. Best performances are marked in bold, and underlined numbers are the best per sample size.}
    \resizebox{\linewidth}{!}{
    \begin{tabular}{@{}lcccccccccccc@{}}
        \toprule 
        & \textbf{Method} & \textbf{0} & \textbf{1} & \textbf{2} & \textbf{3} & \textbf{4} & \textbf{5} & \textbf{6} & \textbf{7} & \textbf{8} & \textbf{9} & \textbf{Average} \\ 
        \midrule
        \multirow{4}{*}{\begin{sideways}1 sample\end{sideways}}
        & Random & 91.11 \std{2.4} & 90.34 \std{6.0} & 65.18 \std{10.6} & 77.02 \std{4.4} & 69.79 \std{11.2} & 69.69 \std{9.7} & 69.43 \std{6.9} & 70.37 \std{7.6} & 76.91 \std{3.6} & 81.26 \std{5.4} & 76.11 \std{2.2} \\ 
        & Greedy & \underline{98.14} & \underline{97.09} & \underline{82.74} & \underline{89.92} & \underline{92.94} & \underline{88.69} & \underline{85.01} & \underline{87.87} & \underline{89.59} & \underline{89.53} & \underline{90.15} \\ 
        & Evo & 82.31 & 83.30 & 69.87 & 75.18 & 78.13 & 82.36 & 78.57 & 59.38 & 62.56 & 76.56 & 74.82 \\ 
        & Core-set & 94.71 & 92.27 & 71.53 & 86.66 & 80.33 & 78.28 & 75.39 & 76.65 & 82.48 & 84.61 & 82.29 \\ 
        \midrule 
        \multirow{4}{*}{\begin{sideways}5 samples\end{sideways}}
        & Random & 98.52 \std{0.9} & 97.79 \std{2.2} & 77.40 \std{7.3} & 89.79 \std{1.9} & 90.25 \std{1.9} & 85.58 \std{4.9} & 88.90 \std{4.3} & 86.97 \std{2.6} & 87.92 \std{3.5} & 90.42 \std{2.1} & 89.36 \std{1.1} \\ 
        & Greedy & 99.21 & 98.29 & 85.59 & 93.13 & 95.65 & 90.90 & 93.00 & \underline{92.52} & \underline{94.31} & 93.56 & \underline{93.62} \\ 
        & Evo & 98.76 & 96.96 & 85.06 & \underline{94.82} & \underline{95.78} & \underline{91.44} & \underline{95.57} & 89.28 & 87.35 & \underline{94.59} & 92.96 \\ 
        & Core-set & \underline{99.29} & \underline{99.58} & \underline{86.83} & 92.32 & 89.07 & 91.30 & 94.72 & 91.68 & 91.88 & 88.82 & 92.55 \\ 
        \midrule 
        \multirow{4}{*}{\begin{sideways}10 samples\end{sideways}}
        & Random & 98.86 \std{0.6} & 99.18 \std{0.4} & 83.04 \std{4.0} & 92.39 \std{2.0} & 93.82 \std{2.3} & 90.84 \std{1.3} & 92.81 \std{2.8} & 93.28 \std{1.2} & 92.34 \std{1.8} & 93.40 \std{1.1} & 93.00 \std{0.6} \\ 
        & Greedy & \underline{99.43} & 98.83 & 89.00 & 93.99 & 96.89 & 91.39 & 95.62 & 93.97 & \underline{96.28} & 93.75 & 94.91 \\ 
        & Evo & 99.40 & 99.30 & 88.08 & \underline{94.42} & \underline{97.97} & \underline{92.40} & \underline{97.59} & \underline{97.28} & 92.17 & 95.09 & \underline{95.37} \\ 
        & Core-set & 99.38 & \underline{99.66} & \underline{90.60} & 94.23 & 96.27 & 91.46 & 95.65 & 93.56 & 93.45 & \underline{95.16} & 94.94 \\ 
        \midrule 
        \multirow{4}{*}{\begin{sideways}25 samples\end{sideways}}
        & Random & 99.38 \std{0.2} & 99.70 \std{0.1} & 90.26 \std{2.1} & 95.36 \std{1.5} & 96.66 \std{1.0} & 93.37 \std{1.0} & 97.55 \std{1.3} & 97.49 \std{0.5} & 95.83 \std{0.6} & 96.79 \std{0.5} & 96.24 \std{0.3} \\ 
        & Greedy & 99.67 & 99.30 & 89.85 & 96.63 & 97.67 & 90.84 & 98.32 & 94.88 & \textbf{\underline{97.86}} & 95.60 & 96.06 \\ 
        & Evo & 99.44 & \underline{99.72} & 91.65 & \underline{97.09} & \textbf{\underline{99.01}} & 95.10 & \textbf{\underline{99.23}} & \underline{98.70} & 95.43 & \underline{97.60} & 97.30 \\ 
        & Core-set & \underline{99.76} & 99.71 & \underline{92.42} & 96.22 & 97.04 & \underline{95.96} & 98.89 & 98.66 & 97.37 & 97.23 & \underline{97.33} \\ 
        \midrule 
        & Full training & \textbf{99.84} & \textbf{99.93} & \textbf{96.58} & \textbf{98.02} & 98.63 & \textbf{96.45} & 99.13 & \textbf{99.14} & 97.82 & \textbf{98.53} & \textbf{98.41} \\ 
        \bottomrule
    \end{tabular}}
    \vspace{2pt}
    \label{tab:results_mnist}
\end{table*}

\begin{table*}
    \centering
    \small
    \caption{Detailed training results for Fashion-MNIST. AUROC scores of full training and training with 1, 5, 10, and 25 best-performing (with greedy search, the evolutionary algorithm, and core-set selection) or random samples. Since the performance of randomly selected subgroups can vary strongly, we repeated these experiments over ten different subsets. Best performances are marked in bold, and underlined numbers are the best per sample size.}
    \resizebox{\linewidth}{!}{
    \begin{tabular}{@{}lcccccccccccc@{}}
        \toprule 
        & \textbf{Method} & \textbf{0} & \textbf{1} & \textbf{2} & \textbf{3} & \textbf{4} & \textbf{5} & \textbf{6} & \textbf{7} & \textbf{8} & \textbf{9} & \textbf{Average} \\ 
        \midrule
        \multirow{4}{*}{\begin{sideways}1 sample\end{sideways}}
        & Random & 77.24 \std{11.0} & 96.45 \std{0.8} & 82.47 \std{17.5} & 73.80 \std{7.5} & 75.09 \std{12.5} & 91.63 \std{4.1} & 69.29 \std{12.4} & 94.71 \std{6.2} & 79.04 \std{8.4} & 95.37 \std{4.6} & 83.51 \std{3.3} \\ 
        & Greedy & \underline{93.85} & \underline{98.12} & \underline{93.03} & \underline{88.85} & \underline{90.46} & \underline{97.14} & \underline{84.55} & \underline{98.84} & \underline{95.99} & \underline{99.16} & \underline{94.00} \\ 
        & Evo & 89.28 & 83.19 & 88.71 & 83.60 & 79.98 & 94.35 & 71.09 & 96.53 & 84.16 & 97.30 & 86.82 \\ 
        & Core-set & 90.94 & 97.12 & 91.63 & 83.53 & 88.62 & 95.75 & 76.01 & 98.67 & 94.42 & 98.00 & 91.47 \\ 
        \midrule
        \multirow{4}{*}{\begin{sideways}5 samples\end{sideways}}
        & Random & 90.58 \std{3.7} & 97.90 \std{0.5} & 92.13 \std{1.8} & 85.24 \std{5.4} & 88.77 \std{1.3} & 95.38 \std{1.9} & 80.89 \std{2.4} & 98.34 \std{0.8} & 91.01 \std{6.7} & 98.41 \std{1.2} & 91.86 \std{1.1} \\ 
        & Greedy & \underline{94.64} & \underline{98.98} & \underline{94.27} & \underline{92.38} & \underline{92.95} & \underline{97.98} & \underline{86.42} & \underline{99.18} & \underline{96.84} & \underline{99.42} & \underline{95.31} \\ 
        & Evo & 92.74 & 98.85 & 93.94 & 89.42 & 88.10 & 94.19 & 83.62 & 98.59 & 96.30 & 98.43 & 93.42 \\ 
        & Core-set & 92.05 & 97.95 & 92.91 & 91.24 & 89.85 & 97.32 & 84.05 & 98.61 & 96.10 & 98.87 & 93.89 \\ 
        \midrule
        \multirow{4}{*}{\begin{sideways}10 samples\end{sideways}}
        & Random & 91.78 \std{2.1} & 98.31 \std{0.5} & 93.22 \std{0.9} & 89.62 \std{2.4} & 89.87 \std{0.7} & 96.24 \std{1.3} & 81.83 \std{1.9} & 98.57 \std{0.7} & 94.31 \std{1.6} & 98.62 \std{0.7} & 93.24 \std{0.3} \\ 
        & Greedy & \underline{94.79} & 98.98 & 94.53 & 93.38 & \underline{92.00} & \textbf{\underline{98.55}} & \underline{86.74} & \underline{99.27} & 96.02 & \underline{99.50} & \underline{95.38} \\ 
        & Evo & 91.41 & \underline{99.31} & \underline{94.57} & \underline{94.72} & 90.11 & 95.62 & 82.86 & 98.93 & \underline{96.90} & 98.37 & 94.28 \\ 
        & Core-set & 93.77 & 98.85 & 94.14 & 92.82 & 91.40 & 96.13 & 84.14 & 98.72 & 95.98 & 98.70 & 94.47 \\ 
        \midrule
        \multirow{4}{*}{\begin{sideways}25 samples\end{sideways}}
        & Random & 93.28 \std{1.2} & 98.62 \std{0.3} & 93.74 \std{0.7} & 93.37 \std{1.6} & 91.48 \std{0.9} & 95.86 \std{1.0} & 82.45 \std{1.4} & 98.89 \std{0.2} & 95.31 \std{1.0} & 98.62 \std{0.5} & 94.16 \std{0.2} \\ 
        & Greedy & \underline{95.06} & 99.03 & \textbf{\underline{94.64}} & 94.49 & \underline{92.97} & \underline{98.54} & \textbf{\underline{86.76}} & \underline{99.26} & 94.86 & \textbf{\underline{99.60}} & \underline{95.52} \\ 
        & Evo & 94.10 & \underline{99.24} & 94.12 & \underline{95.15} & \underline{92.97} & 95.70 & 84.11 & 98.94 & 97.32 & 99.00 & 95.07 \\ 
        & Core-set & 94.40 & 99.18 & 94.16 & 94.70 & 91.64 & 97.07 & 84.80 & 99.13 & \underline{97.37} & 98.87 & 95.13 \\ 
        \midrule 
        & Full training & \textbf{95.09} & \textbf{99.52} & 94.44 & \textbf{96.28} & \textbf{93.66} & 96.58 & 85.33 & \textbf{99.28} & \textbf{98.88} & 98.86 & \textbf{95.79} \\ 
        \bottomrule
    \end{tabular}}
    \label{tab:results_fashion_mnist}
\end{table*}

\begin{table*}
    \centering
    \small
    \caption{Detailed training results for MVTec-AD. AUROC scores of full training and training with 1, 5, 10, and 25 best-performing (with greedy search and evolutionary algorithm) or random samples. Since the performance of randomly selected subgroups can vary strongly, we repeated these experiments over ten different subsets. Best performances are marked in bold, and underlined numbers are the best per sample size.}
    \resizebox{\linewidth}{!}{
    \begin{tabular}{@{}lccccccccc@{}}
        \toprule 
        & \textbf{Method} & \textbf{Bottle} & \textbf{Cable} & \textbf{Capsule} & \textbf{Carpet} & \textbf{Grid} & \textbf{Hazelnut} & \textbf{Leather} & \textbf{Metal nut} \\ 
        \midrule 
        \multirow{4}{*}{\begin{sideways}1 sample\end{sideways}}
        & Random & 99.71 \std{0.1} & 83.74 \std{4.7} & 66.80 \std{5.1} & 97.72 \std{0.5} & 60.30 \std{6.9} & 90.67 \std{2.6} & 99.99 \std{0.0} & 71.99 \std{4.0} \\ 
        & Greedy & \underline{99.76} & 88.19 & \underline{72.80} & 98.60 & \underline{71.19} & \underline{93.93} & \textbf{\underline{100.00}} & \underline{72.19} \\ 
        & Evo & 99.52 & 88.92 & 63.14 & 98.23 & 66.88 & 93.04 & \textbf{\underline{100.00}} & 70.97 \\ 
        & Core-set & 99.52 & \underline{89.81} & 61.55 & \textbf{\underline{99.08}} & 65.99 & 88.68 & \textbf{\underline{100.00}} & 71.07 \\ 
        \midrule 
        \multirow{4}{*}{\begin{sideways}5 samples\end{sideways}}
        & Random & 99.84 \std{0.2} & 91.40 \std{3.6} & 84.11 \std{7.2} & 97.98 \std{0.3} & 72.64 \std{7.1} & 96.29 \std{2.2} & \textbf{100.00} \std{0.0} & 94.93 \std{3.5} \\ 
        & Greedy & \textbf{\underline{100.00}} & \underline{96.27} & 84.96 & \underline{98.64} & 73.52 & 98.29 & \textbf{\underline{100.00}} & 97.75 \\ 
        & Evo & 99.52 & 93.22 & \underline{90.91} & 98.19 & \underline{86.04} & \underline{99.14} & \textbf{\underline{100.00}} & \underline{98.34} \\ 
        & Core-set & 99.68 & 92.82 & 87.36 & 98.72 & 69.40 & 98.64 & \textbf{\underline{100.00}} & 96.82 \\ 
        \midrule 
        \multirow{4}{*}{\begin{sideways}10 samples\end{sideways}}
        & Random & 99.90 \std{0.2} & 93.49 \std{1.8} & 90.29 \std{2.3} & 98.04 \std{0.3} & 80.91 \std{5.7} & 98.99 \std{0.7} & \textbf{100.00} \std{0.0} & 97.72 \std{1.6} \\ 
        & Greedy & \textbf{\underline{100.00}} & \underline{97.58} & \underline{93.06} & 98.15 & 78.86 & 99.64 & \textbf{\underline{100.00}} & 98.92 \\ 
        & Evo & \textbf{\underline{100.00}} & 96.42 & 91.26 & 98.39 & \underline{94.40} & \textbf{\underline{100.00}} & \textbf{\underline{100.00}} & \underline{99.07} \\ 
        & Core-set & 99.60 & 92.75 & 90.75 & \underline{98.52} & 76.38 & 99.68 & \textbf{\underline{100.00}} & 98.19 \\ 
        \midrule 
        \multirow{4}{*}{\begin{sideways}25 samples\end{sideways}}
        & Random & \textbf{\underline{100.00}} \std{0.0} & 96.59 \std{1.0} & 93.61 \std{1.5} & 98.14 \std{0.3} & 90.23 \std{2.9} & 99.81 \std{0.3} & \textbf{100.00} \std{0.0} & 99.20 \std{0.4} \\
        & Greedy & \textbf{\underline{100.00}} & 96.74 & 93.94 & 98.27 & 83.88 & 99.71 & \textbf{\underline{100.00}} & 99.41 \\ 
        & Evo & \textbf{\underline{100.00}} & \underline{98.29} & \underline{94.34} & 98.56 & \textbf{\underline{99.11}} & \textbf{\underline{100.00}} & \textbf{\underline{100.00}} & \underline{99.76} \\ 
        & Core-set & 99.68 & 97.49 & 91.58 & \underline{98.88} & 92.86 & \textbf{\underline{100.00}} & \textbf{\underline{100.00}} & 99.46 \\ 
        \midrule 
        & Full training & \textbf{100.00} & \textbf{99.53} & \textbf{99.20} & 98.43 & 99.08 & \textbf{100.00} & \textbf{100.00} & \textbf{99.90} \\ 
        \midrule
        \midrule
        & \textbf{Method} & \textbf{Pill} & \textbf{Screw} & \textbf{Tile} & \textbf{Toothbrush} & \textbf{Transistor} & \textbf{Wood} & \textbf{Zipper} & \textbf{Average} \\ 
        \midrule
        \multirow{4}{*}{\begin{sideways}1 sample\end{sideways}}
        & Random & 78.71 \std{5.0} & 46.22 \std{4.8} & 99.59 \std{0.5} & 82.58 \std{3.5} & 83.31 \std{4.7} & 98.18 \std{0.7} & 94.58 \std{1.5} & 83.61 \std{1.2} \\ 
        & Greedy & \underline{89.23} & \underline{55.87} & \textbf{\underline{100.00}} & 88.61 & 91.42 & \underline{99.21} & \underline{99.37} & \underline{89.70} \\ 
        & Evo & 72.64 & 52.96 & 99.13 & \underline{90.00} & \underline{92.12} & 99.04 & 96.61 & 85.55 \\ 
        & Core-set & 74.58 & 38.29 & 98.63 & 85.83 & 82.79 & \underline{99.21} & 95.64 & 82.80 \\ 
        \midrule
        \multirow{4}{*}{\begin{sideways}5 samples\end{sideways}}
        & Random & 89.48 \std{2.0} & 52.86 \std{5.1} & 99.87 \std{0.1} & 87.22 \std{5.2} & 93.77 \std{1.7} & 98.57 \std{0.4} & 97.55 \std{1.5} & 90.43 \std{0.9} \\ 
        & Greedy & 89.77 & 53.40 & \textbf{\underline{100.00}} & 87.22 & 94.46 & \underline{99.30} & \underline{99.16} & 91.52 \\ 
        & Evo & \underline{91.41} & \underline{61.47} & 99.49 & 98.33 & \underline{97.58} & \underline{99.30} & 98.90 & \underline{94.12} \\ 
        & Core-set & 85.00 & 52.82 & 99.57 & \underline{98.89} & 96.50 & 98.77 & 95.93 & 92.13 \\ 
        \midrule
        \multirow{4}{*}{\begin{sideways}10 samples\end{sideways}}
        & Random & 90.95 \std{1.8} & 58.42 \std{4.0} & 99.89 \std{0.1} & 90.75 \std{1.1} & 95.86 \std{1.8} & 98.60 \std{0.2} & 98.13 \std{1.0} & 92.80 \std{0.6} \\ 
        & Greedy & \underline{93.43} & 53.68 & \textbf{\underline{100.00}} & 85.56 & 96.38 & 99.30 & \textbf{\underline{99.58}} & 92.94 \\ 
        & Evo & 93.40 & \underline{75.90} & 99.71 & 99.44 & \underline{99.58} & \underline{99.39} & 98.58 & \underline{96.37} \\ 
        & Core-set & 89.20 & 61.26 & 98.85 & \textbf{\underline{100.00}} & 99.17 & 98.51 & 96.77 & 95.04 \\ 
        \midrule
        \multirow{4}{*}{\begin{sideways}25 samples\end{sideways}}
        & Random & 93.73 \std{1.1} & 72.89 \std{4.0} & 99.94 \std{0.1} & 90.06 \std{0.6} & 97.88 \std{0.7} & 98.62 \std{0.2} & 98.64 \std{0.6} & 95.29 \std{0.3} \\ 
        & Greedy & 94.03 & 54.44 & \textbf{\underline{100.00}} & 85.56 & 98.17 & 99.21 & \underline{99.55} & 93.53 \\ 
        & Evo & \underline{95.23} & \underline{95.43} & 98.77 & 99.72 & 99.50 & \textbf{\underline{99.56}} & \underline{99.55} & \textbf{\underline{98.52}} \\ 
        & Core-set & 94.38 & 89.44 & 99.57 & \textbf{\underline{100.00}} & \underline{99.58} & 98.68 & 98.37 & 96.80 \\ 
        \midrule 
        & Full training & \textbf{96.21} & \textbf{97.13} & 99.96 & 90.28 & \textbf{99.62} & 98.77 & 99.11 & 98.48 \\ 
        \bottomrule
    \end{tabular}}
    \label{tab:results_mvtec}
\end{table*}

\begin{table*}
    \centering
    \small
    \caption{Detailed training results for CIFAR100. AUROC scores of full training and training with 1, 5, 10, and 25 best-performing (with greedy search, the evolutionary algorithm, and core-set selection) or random samples. Since the performance of randomly selected subgroups can vary strongly, we repeated these experiments over ten different subsets. Best performances are marked in bold, and underlined numbers are the best per sample size.}
    \resizebox{\linewidth}{!}{
    \begin{tabular}{@{}lccccccccccc@{}}
        \toprule 
        & \thead{Method} & \thead{Aquatic \\ mammals} & \thead{Fish} & \thead{Flowers} & \thead{Food \\ containers} & \thead{Fruit and \\ vegetables} & \thead{Household \\ electrical \\ devices} & \thead{Household \\ furniture} & \thead{Insects} & \thead{Large \\ carnivores} & \thead{Large \\ man-made \\ outdoor things} \\ 
        \midrule
        \multirow{4}{*}{\begin{sideways}1 sample\end{sideways}}
        & Random & 71.70 \std{14.7} & 75.23 \std{8.1} & 90.83 \std{3.9} & 82.25 \std{3.7} & 71.29 \std{12.5} & 71.88 \std{7.1} & 84.70 \std{6.1} & 68.09 \std{12.3} & 77.95 \std{11.7} & 86.72 \std{6.0} \\ 
        & Greedy & \underline{91.93} & \underline{89.09} & \underline{96.78} & \underline{89.39} & \underline{90.97} & \underline{85.73} & \underline{96.17} & \underline{84.78} & \underline{92.46} & \underline{93.98} \\ 
        & Evo & 89.68 & 78.96 & 80.31 & 76.69 & 84.76 & 80.61 & 92.09 & 63.07 & 85.80 & 90.19 \\ 
        & Core-set & 64.34 & 75.71 & 90.52 & 80.05 & 85.14 & 73.08 & 90.44 & 61.80 & 68.79 & 89.97 \\ 
        \midrule
        \multirow{4}{*}{\begin{sideways}5 samples\end{sideways}}
        & Random & 86.25 \std{3.8} & 81.82 \std{6.1} & 95.82 \std{1.4} & 89.31 \std{2.4} & 85.26 \std{5.9} & 78.90 \std{6.2} & 92.54 \std{2.0} & 79.26 \std{9.0} & 87.47 \std{1.7} & 92.15 \std{1.5} \\ 
        & Greedy & \textbf{\underline{94.13}} & \underline{92.32} & \underline{98.13} & \underline{95.59} & \underline{93.00} & 85.05 & \underline{96.78} & \underline{91.05} & \underline{93.73} & \underline{95.55} \\ 
        & Evo & 93.04 & 89.85 & 97.08 & 85.89 & 92.47 & \underline{89.12} & 95.64 & 86.96 & 90.11 & 93.78 \\ 
        & Core-set & 86.53 & 88.90 & 96.48 & 91.79 & 91.29 & 86.67 & 95.75 & 87.01 & 91.77 & 92.95 \\ 
        \midrule
        \multirow{4}{*}{\begin{sideways}10 samples\end{sideways}}
        & Random & 88.50 \std{1.9} & 85.56 \std{4.8} & 96.97 \std{0.7} & 91.31 \std{1.7} & 90.59 \std{3.1} & 80.97 \std{4.3} & 94.77 \std{0.8} & 81.96 \std{5.0} & 90.22 \std{1.4} & 92.49 \std{1.2} \\ 
        & Greedy & \underline{93.84} & \underline{91.92} & \textbf{\underline{98.60}} & \textbf{\underline{95.97}} & 92.88 & 82.56 & 96.81 & \underline{89.68} & \underline{93.33} & \textbf{\underline{95.57}} \\ 
        & Evo & 93.19 & 89.98 & 98.16 & 91.44 & \underline{93.24} & \underline{88.27} & \underline{96.86} & 88.92 & 90.95 & 93.79 \\ 
        & Core-set & 91.31 & 90.85 & 97.41 & 92.08 & 92.28 & 86.71 & 95.99 & 85.73 & 92.60 & 94.00 \\ 
        \midrule
        \multirow{4}{*}{\begin{sideways}25 samples\end{sideways}}
        & Random & 90.04 \std{1.3} & 90.78 \std{1.2} & 97.87 \std{0.5} & 93.19 \std{1.5} & 93.63 \std{0.6} & 84.21 \std{4.1} & 96.28 \std{0.5} & 86.34 \std{1.7} & 91.82 \std{0.8} & 93.50 \std{0.6} \\ 
        & Greedy & 92.36 & 91.27 & \underline{98.50} & \underline{95.79} & 92.53 & 80.51 & \textbf{\underline{97.26}} & \underline{90.89} & \underline{93.21} & \underline{95.45} \\ 
        & Evo & \underline{92.88} & \underline{92.50} & 98.39 & 92.87 & \underline{95.00} & 90.37 & 96.84 & 90.74 & 91.84 & 94.29 \\ 
        & Core-set & 91.72 & 91.87 & 98.23 & 94.68 & 94.94 & \underline{90.80} & 96.21 & 90.31 & 93.09 & 93.92 \\ 
        \midrule 
        & Full training & 93.68 & \textbf{94.81} & 98.46 & 95.86 & \textbf{96.59} & \textbf{94.59} & \textbf{97.26} & \textbf{93.00} & \textbf{95.34} & 94.98 \\ 
        \midrule
        \midrule
        & \thead{Method} & \thead{Large natural \\ outdoor scenes} & \thead{Large \\ omnivores \\ and herbivores} & \thead{Medium-sized \\ mammals} & \thead{Non-insect \\ invertebrates} & \thead{People} & \thead{Reptiles} & \thead{Small \\ mammals} & \thead{Trees} & \thead{Vehicles 1} & \thead{Vehicles 2} \\ 
        \midrule 
        \multirow{4}{*}{\begin{sideways}1 sample\end{sideways}}
        & Random & 86.87 \std{6.3} & 66.79 \std{14.8} & 73.91 \std{7.6} & 60.08 \std{8.3} & 88.73 \std{6.4} & 66.99 \std{7.5} & 72.46 \std{9.3} & 92.30 \std{3.6} & 78.59 \std{8.2} & 73.74 \std{6.6} \\ 
        & Greedy & \underline{94.13} & \underline{86.01} & \underline{84.57} & \underline{79.51} & \underline{96.20} & \underline{79.39} & \underline{88.86} & \underline{95.92} & \underline{93.20} & \underline{89.81} \\ 
        & Evo & 88.88 & 78.12 & 73.68 & 77.34 & 83.52 & 52.97 & 86.70 & 95.58 & 68.73 & 72.15 \\ 
        & Core-set & 93.57 & 74.96 & 79.55 & 56.37 & 91.10 & 72.36 & 87.23 & 94.28 & 82.56 & 80.49 \\ 
        \midrule 
        \multirow{4}{*}{\begin{sideways}5 samples\end{sideways}}
        & Random & 92.30 \std{2.0} & 82.54 \std{3.3} & 84.66 \std{2.1} & 71.37 \std{5.1} & 95.66 \std{1.2} & 79.22 \std{3.2} & 85.18 \std{4.6} & 95.67 \std{0.9} & 88.96 \std{3.2} & 83.75 \std{3.8} \\ 
        & Greedy & 94.98 & \underline{90.29} & \underline{89.08} & \underline{82.67} & \underline{97.90} & 84.47 & \underline{92.49} & \underline{96.89} & \underline{95.61} & \underline{90.10} \\ 
        & Evo & \underline{95.11} & 87.96 & 83.67 & 81.76 & 97.33 & 81.79 & 90.32 & 96.83 & 92.12 & 87.22 \\ 
        & Core-set & 93.92 & 86.92 & 87.56 & 75.77 & 95.14 & \underline{85.19} & 86.94 & 95.48 & 92.46 & 86.36 \\ 
        \midrule 
        \multirow{4}{*}{\begin{sideways}10 samples\end{sideways}}
        & Random & 93.76 \std{1.2} & 85.27 \std{2.7} & 87.79 \std{1.3} & 78.13 \std{2.9} & 96.39 \std{0.8} & 83.14 \std{3.0} & 87.63 \std{2.7} & 96.44 \std{0.5} & 92.30 \std{1.8} & 86.80 \std{1.5} \\ 
        & Greedy & 95.49 & 89.41 & \underline{90.43} & \underline{85.01} & \underline{98.18} & \underline{87.55} & \underline{91.98} & \underline{97.67} & \underline{95.92} & 88.95 \\ 
        & Evo & 94.95 & \underline{89.90} & 88.72 & 83.32 & 97.69 & 85.42 & 90.41 & 97.48 & 94.83 & 88.14 \\ 
        & Core-set & \underline{95.81} & 86.39 & 87.80 & 79.06 & 96.93 & 86.52 & 89.47 & 96.81 & 93.88 & \underline{89.05} \\ 
        \midrule 
        \multirow{4}{*}{\begin{sideways}25 samples\end{sideways}}
        & Random & 94.93 \std{0.7} & 89.69 \std{1.5} & 89.68 \std{0.6} & 83.07 \std{1.6} & 97.47 \std{0.5} & 86.95 \std{1.1} & 89.92 \std{0.8} & 97.06 \std{0.3} & 94.09 \std{0.4} & 89.90 \std{1.0} \\ 
        & Greedy & \textbf{\underline{95.84}} & 90.23 & 90.78 & \underline{86.55} & \underline{98.13} & \underline{89.72} & \underline{91.81} & 97.50 & \underline{95.87} & 87.78 \\ 
        & Evo & 95.44 & 91.58 & \underline{91.27} & 85.68 & 97.97 & 88.61 & 91.33 & 97.17 & 95.64 & 90.17 \\ 
        & Core-set & 95.81 & \underline{91.72} & 90.12 & 83.86 & 97.56 & 88.04 & 90.91 & \textbf{\underline{97.84}} & 94.60 & \underline{91.28} \\ 
        \midrule 
        & Full training & 95.37 & \textbf{94.40} & \textbf{93.54} & \textbf{90.76} & \textbf{98.55} & \textbf{91.71} & \textbf{92.70} & 96.84 & \textbf{96.06} & \textbf{93.88} \\ 
        \bottomrule
    \end{tabular}}
    \label{tab:results_cifar100}
\end{table*}

\end{document}